\documentclass[sn-mathphys-num]{sn-jnl}

\usepackage{graphicx}%
\usepackage{amsmath,amssymb,amsfonts}%
\usepackage{amsthm}%
\usepackage{pifont}
\usepackage{mathrsfs}%
\usepackage[title]{appendix}%
\usepackage{scontents}%
\usepackage{array} %
\usepackage{nameref}
\usepackage{makecell}

\usepackage{booktabs}
\usepackage{threeparttable}
\usepackage{multirow}
\usepackage{rotating}
\usepackage{float}
\usepackage{subcaption}
\usepackage{comment}

\newgeometry{left=2cm,right=2cm, top=2cm, bottom=2cm}   

\usepackage{lineno}
\linenumbers
\theoremstyle{thmstyleone}%
\theoremstyle{thmstyletwo}%

\theoremstyle{thmstylethree}%

\begin{document}

\title[Large Language Model–Based Generation of Discharge Summaries]{Large Language Model–Based Generation of Discharge Summaries}


\author*[1,2]{\fnm{Tiago} \sur{Rodrigues}}\email{up201907021@edu.fe.up.pt}

\author[1,2]{\fnm{Carla} \sur{Teixeira} \sur{Lopes}}

\affil[1]{\orgdiv{Faculty of Engineering}, \orgname{University of Porto}, \orgaddress{\city{Porto}, \country{Portugal}}}

\affil[2]{\orgname{INESC TEC}, \orgaddress{\city{Porto}, \country{Portugal}}}



\abstract{Discharge Summaries are documents written by medical professionals that detail a patient’s visit to a care facility. They contain a wealth of information crucial for patient care, and automating their generation could significantly reduce the effort required from healthcare professionals, minimize errors, and ensure that critical patient information is easily accessible and actionable. In this work, we explore the use of five Large Language Models on this task, from open-source models (Mistral, Llama 2) to proprietary systems (GPT-3, GPT-4, Gemini 1.5 Pro), leveraging MIMIC-III summaries and notes. We evaluate them using exact-match, soft-overlap, and reference-free metrics. Our results show that proprietary models, particularly Gemini with one-shot prompting, outperformed others, producing summaries with the highest similarity to the gold-standard ones. Open-source models, while promising, especially Mistral after fine-tuning, lagged in performance, often struggling with hallucinations and repeated information. Human evaluation by a clinical expert confirmed the practical utility of the summaries generated by proprietary models. Despite the challenges, such as hallucinations and missing information, the findings suggest that LLMs, especially proprietary models, are promising candidates for automatic discharge summary generation as long as data privacy is ensured.}

\nolinenumbers
\maketitle
\addcontentsline{toc}{section}{\protect\numberline{}Abstract}

\section{Introduction}
\label{sec:introduction}

    When a patient is admitted to a hospital, critical information is documented in clinical notes. These notes are crucial for tracking a patient's issues and progress and facilitate information sharing among medical professionals. Upon discharge, these notes are compiled into a discharge summary, a comprehensive document that includes all the essential details captured during a patient's admission, his hospital course, final diagnosis, and follow-up care~\cite{wimsettReviewArticleComponents2014}. This process, however, is cumbersome and time-intensive, requiring clinicians to invest significant amounts of time analyzing and summarizing vast amounts of information, an activity which can be prone to errors or misjudgment~\cite{searleDischargeSummaryHospital2023}. This workload adds to the already high demands on healthcare providers, contributing to exhaustion and reducing the time available for direct patient care~\cite{westPhysicianBurnoutContributors2018,shingClinicalEncounterSummarization2021}.
    
    Fortunately, in recent years, advancements in Large Language Models (LLMs) have made them highly competitive in a myriad of Natural Language Processing tasks, including Information Extraction, Text Generation, and Summarization~\cite{naveedComprehensiveOverviewLarge2024}. These advancements have made them competent in the problems clinicians face, making it possible to automate this process. While limited context windows previously restricted their applicability, recent models such as GPT-4~\cite{openaiGPT4TechnicalReport2023} and Google's Gemini 1.5 Pro~\cite{geminiteamGeminiFamilyHighly2024} improved their limits significantly, enabling the processing of entire sets of notes in one go.
    
    This work explores the ability of five recently developed LLMs to generate a discharge summary from the set of notes written during a patient's admission. We cover two main classes of LLMs: smaller yet powerful open-source models:  Mistral-7B~\cite{jiangMistral7B2023}, and Llama2-7B~\cite{touvronLlamaOpenFoundation2023}; and more powerful but proprietary models: GPT-3.5, GPT-4, and Gemini 1.5 Pro. We experimented with two main techniques for improving the results, fine-tuning and one-shot prompting, and conducted an extensive analysis of their capabilities. Proprietary models show overall superior performance, but the fine-tuned version of Mistral produces competitive results. Still, significant advancements are needed for real-world applicability. We release the source code used to conduct the experiments
    \footnote{Available on \href{https://github.com/1Krypt0/clinical-summarization-llm}{https://github.com/1Krypt0/clinical-summarization-llm.}}.

\section{Related Work}
\label{sec:background}
    
    Summarization is a Natural Language Processing (NLP) task that aims to distill key information from a larger body of text by preserving its essential content. Approaches can be \emph{extractive}, selecting key sentences from the original text, or \emph{abstractive}, generating new sentences that convey the original meaning in a condensed format~\cite{jangraSurveyMultimodalSummarization2023,kohEmpiricalSurveyLong2022,maMultidocumentSummarizationDeep2023,yadavStateoftheartApproachExtractive2023a}. This task can also be classified as either single-document or multi-document summarization~\cite{maMultidocumentSummarizationDeep2023}, depending on the data source.

    Summarization has a growing prevalence in the clinical field~\cite{rodriguesHarnessingLargeLanguage2025}, where it has been applied to medical evidence from research articles~\cite{tangEvaluatingLargeLanguage2023}, synthesizing a discharge report into a discharge summary~\cite{zhuLeveragingSummaryGuidance2023a}, and generating brief hospital course sections of discharge summaries~\cite{adams-etal-2021-whats}. Popular models include decoder-only architectures such as GPT-3.5~\cite{tangEvaluatingLargeLanguage2023}, and seq2seq models like BART~\cite{alaeiAutomatedDischargeSummary2023,searleDischargeSummaryHospital2023,zhuLeveragingSummaryGuidance2023a}, T5~\cite{zhuLeveragingSummaryGuidance2023a,searleDischargeSummaryHospital2023}, and BERT2BERT~\cite{zhuLeveragingSummaryGuidance2023a,searleDischargeSummaryHospital2023}.

    Discharge summary generation has been a topic of rising research interest~\cite{luLargeLanguageModels2024}, but it doesn't yet have a significant body of work~\cite{maMultidocumentSummarizationDeep2023}. An article by~\citet{adams-etal-2021-whats} established in 2021 the task of hospital-course summarization and identified important features to be considered when approaching it.  In the same year, one of the first works generating discharge summaries from prior notes was published by Shing et al.~\cite{shingClinicalEncounterSummarization2021}. More recently, the shared task by Xu et al.~\cite{xu-etal-2024-overview} extended this line of work with new developments. Some attempts to automatically generate a discharge summary focused on specific parts of the report, such as the Chief Complaint~\cite{leeNaturalLanguageGeneration2018} or the Brief Hospital Course~\cite{searleDischargeSummaryHospital2023}. 

    Simplifying and summarizing the information on discharge reports is a task that has also gained traction~\cite{kimPatientFriendly,GOSWAMI2024111531}. For example, the work of Goswami et al.~\cite{GOSWAMI2024111531} is a recent attempt to tackle the summarization of the entire discharge summary, leveraging Llama-2 to perform the task. They attempt to summarize 700 discharge records, creating condensed and easier-to-understand reports for patients. For this, they first generate reference summaries for each report and train the model with different techniques, of which QLoRA stands as the most effective. While the task is similar to ours, it serves a different purpose and does not tackle multi-document summarization, as they leverage the already compiled report instead of the collection of notes.

    A paper that works on the same objective is that of Ellershaw et al.~\cite{ellershaw2024automated}, who propose a system leveraging GPT-4 Turbo for this task. Additionally, they also used MIMIC-III's data to conduct the experiments and presented the entire set of notes to the model at once. However, their evaluation was only qualitative and conducted on 53 summaries, a figure that makes it challenging to generalize the results. Moreover, their tests were only conducted on GPT-4 Turbo using few-shot prompting and a structured output, which complicates the assessment of how much of the scores are due to these restrictions over the model's inherent capability.
    
    In general, approaches revolve around two methods: Purely abstractive methods~\cite{searleDischargeSummaryHospital2023,alaeiAutomatedDischargeSummary2023,palSummarizationGenerationDischarge,kimPatientFriendly,GOSWAMI2024111531,ellershaw2024automated}, where a single model processes the input and generates a discharge summary, and Extract-then-Abstract methods~\cite{searleDischargeSummaryHospital2023,shingClinicalEncounterSummarization2021}, where key sentences are first selected by an ``extractor'' model, and then summarized by an ``abstractor''.

    Both approaches show promise, although the extract-then-abstract workflow tends to have issues with sentence selection besides inheriting the same problems of the purely abstractive approach, such as repeating words and sentences and generating inconsistencies between the reported information and the true summary~\cite{searleDischargeSummaryHospital2023,shingClinicalEncounterSummarization2021}. Moreover, the sentences tend to be less coherent, and the abstraction models require a deeper understanding of the language to interpret the disconnected sentences~\cite{searleDischargeSummaryHospital2023}.

    One advantage of extract-then-abstract systems is the easier handling of long text sequences, an issue common with multi-document summarization~\cite{searleDischargeSummaryHospital2023,shingClinicalEncounterSummarization2021,maMultidocumentSummarizationDeep2023}. As the extractor only selects meaningful sentences, it reduces the size of the document to an appropriate context length. Authors opting for the purely abstractive approach either try to apply a model to the full sequence~\cite{alaeiAutomatedDischargeSummary2023} risking not being able to process it correctly or crop irrelevant information from the source to generate the summary~\cite{palSummarizationGenerationDischarge}, possibly missing critical information from the clipped sources.
\section{Methodology}
\label{sec:methodology}

    Our goal is to explore a set of LLMs for discharge summary generation. We begin by giving an overview of the models we focused on and how they were adapted to our specific domain. Then, we detail the data used in our experiments. Finally, we describe the metrics used to assess performance.

    \subsection{Model Selection}

        We evaluated two main classes of models: smaller, open-source models and larger, proprietary ones. All models are decoder-only and have promising results in various NLP tasks~\cite{jiangMistral7B2023,touvronLLaMAOpenEfficient2023,brownLanguageModelsAre2020,openaiGPT4TechnicalReport2023,geminiteamGeminiUnlockingMultimodal2024}.

        \subsubsection{Open-Source Models}

            Two criteria were prioritized to select suitable open-source models: model size and context window. Too large models that would not fit on our NVIDIA RTX 3090 (24GB of vRAM) were excluded, and we required a context window sufficiently large to accommodate our data, which we fixed at a minimum of 8,192 tokens.
        
            We leveraged models from the Llama-2~\cite{touvronLLaMAOpenEfficient2023} and Mistral~\cite{jiangMistral7B2023} families. Llama-2 models come in 3 sizes: 7B, 13B, and 70B parameters. Even though they have a context window of 4,096 tokens, they can be extended to our 8,192 token limit due to Rotary Position Embeddings (RoPE)~\cite{suRoFormerEnhancedTransformer2023}, enabling an increase in context length with additional training. Due to the hardware restrictions mentioned above, we opted for the 7B version.
        
            Mistral-7B is a model developed by Mistral AI that reportedly outperforms the bigger Llama-2-13B on various tasks at a much smaller size and faster inference speed~\cite{jiangMistral7B2023}. Moreover, it natively supports 8,192 tokens, so no adaptation is needed.

            We utilized the Instruction-Tuned version of both models to optimize their performance on this task. The specific models we used are available as `mistralai/Mistral-7B-Instruct-v0.2'\footnote{\href{https://huggingface.co/mistralai/Mistral-7B-Instruct-v0.2}{https://huggingface.co/mistralai/Mistral-7B-Instruct-v0.2}} and as `meta-llama/Llama-2-7b-chat-hf'\footnote{\href{https://huggingface.co/meta-llama/Llama-2-7b-chat-hf}{https://huggingface.co/meta-llama/Llama-2-7b-chat-hf}}.

        \subsubsection{Proprietary Models}

            For proprietary language models, we opted to include OpenAI's GPT-3.5~\cite{brownLanguageModelsAre2020}, GPT-4~\cite{openaiGPT4TechnicalReport2023}, and Google's Gemini 1.5 Pro~\cite{geminiteamGeminiUnlockingMultimodal2024}. GPT-4 has been widely regarded as the state-of-the-art in various NLP tasks~\cite{openaiGPT4TechnicalReport2023}, while Gemini 1.5 Pro, an improvement over the previous Gemini 1.0 Pro, appears to have better performance than even the much larger 1.0 Ultra~\cite{geminiteamGeminiUnlockingMultimodal2024}, which in itself showed better performance than GPT-4 in many tasks~\cite{geminiteamGeminiFamilyHighly2024}. These models tend to have much larger context lengths than the open-source ones, with GPT-3.5 having 16,384 tokens, GPT-4 increasing this figure to 128,000 tokens, and Gemini 1.5 Pro to 1 million tokens, allowing for large collections of text to be processed simultaneously.

        \subsection{Model Adaptation}
    \label{subsec:model-adaptation}

        Although the base models have good results in various NLP tasks, one of the standout features of open-source models is their adaptability through fine-tuning. Taking advantage of this potential, we adapted our models to the clinical domain.
    
        For the open-source models, we employed Quantized Low-Rank Adaptation (QLoRA)~\cite{dettmersQLoRAEfficientFinetuning2023}, a resource-efficient variation of LoRA~\cite{huLoRALowRankAdaptation2021}. QLoRA reduces memory requirements by applying a memory-efficient data type, 4-bit NormalFloat (NF4), and Double Quantization, enabling the training of large models using fewer resources~\cite{dettmersQLoRAEfficientFinetuning2023}. To implement it, we utilized torchtune\footnote{\href{https://github.com/pytorch/torchtune}{https://github.com/pytorch/torchtune}}, a PyTorch~\cite{paszkePyTorchImperativeStyle2019a} library for fine-tuning and experimenting with LLMs. Minor modifications were made to the library's source code to accommodate our preformatted instruction templates without additional text alterations.

        Both models were fine-tuned using the AdamW optimizer~\cite{loshchilovDecoupledWeightDecay2019} with a learning rate of $2e^{-5}$. We used a cosine learning rate scheduler with 100 warmup steps and four gradient accumulation steps. Due to memory restrictions, we only applied 1 item per batch.

        The training outcomes varied between models. While Mistral's loss showed only slight improvement (hovering around 0.8 after training), its starting point was already promising. Llama-2 exhibited substantial progress, with loss values decreasing from over 4.0 to around 1.15. Training durations were similar: Mistral required 3 days, 19 hours, and 53 minutes, while Llama-2 took 3 days, 16 hours, and 39 minutes.

        We could not apply QLoRA to the proprietary models because these do not have their weights publicly available. Instead, we opted for one-shot prompting to guide their generation, a proven technique for improving performance~\cite{wangPromptEngineeringHealthcare2024}, as it reduces the dependency on model pre-training and guides the model to a more well-defined answer.

        Except for GPT-3.5, each proprietary model was tested in zero and one-shot scenarios. We opted not to include GPT-3.5 in the one-shot generation because its context window, despite expansive, is not large enough to accommodate more than a single set of notes consistently. For the one-shot scenario, we used the same notes and discharge summary in all tests obtained from a random admission from the training data. To guide the model's response, we asked it to output it in a structured format. 
        
        As a starting point, we devised the prompt format for the zero-shot prompt, of which a snippet is given below. For a full version, refer to the accompanying repository\footnote{\href{https://github.com/1Krypt0/clinical-summarization-llm}{https://github.com/1Krypt0/clinical-summarization-llm}}. The headers used in this formatted prompt were obtained using regular expressions over the entire dataset and are the most prominent ones across all summaries, encapsulating all the relevant details of a discharge summary.
        
        {\footnotesize
        \begin{verbatim}
You are an expert clinical assistant. You will receive a collection of clinical notes. You will summarize them 
in the style of a discharge summary, outputting the following fields, and no additional information:

Admission Date: [Admission Date]
Discharge Date: [Discharge Date]

Date of Birth: [Date of Birth]
Sex: [Sex]

Service: [Hospital Service used]

Chief Complaint: [Chief Complaint]

...
    \end{verbatim}
    }

















        For the one-shot variant, after supplying the text above, we append the following:

        {\footnotesize
        \begin{verbatim}
For example, given the notes:

### NOTES START ###
{notes content}
### NOTES END ###

You should summarize them as:

### SUMMARY START ###
{associated summary content}
### SUMMARY END ###
        \end{verbatim}
        }
        
        We use the titles ``\verb|NOTES START|'' and ``\verb|SUMMARY START|'' to highlight the position of the notes to the model and their respective ends. Also, to prompt the open-source models, we used the following simplified version of the above prompt:
        
        {\footnotesize
        \begin{verbatim}
You are an expert clinical assistant. You will receive a collection of clinical notes. You will summarize them 
in the style of a discharge summary.
        \end{verbatim}
        }
        
        We passed only this sentence, which we called the system prompt, to avoid exceeding the token limits. An overview of all the models used, as well as their potential for adaptation, is given in Table~\ref{tab:models-used}.

        \begin{table*}[ht]
            \centering
            \caption{Models selected for experimentation}
            \label{tab:models-used}
            \begin{threeparttable}
                \centering
                \begin{tabular}{l c c c c l}
                \toprule
                \textbf{Model} & \textbf{Parameters} & \textbf{Context (tokens)} & \textbf{Proprietary?} & \textbf{Trainable?} & \textbf{Prompt Types} \\ \midrule
                Llama-2 & 7B & 4,096\tnote{1} & No & Yes & System Prompt Only \\
                Mistral & 7B & 8,192 & No & Yes & System Prompt Only \\
                GPT-3 & 175B & 16,384 & Yes & No & Zero-Shot \\
                GPT-4 & Unknown & 128,000 & Yes & No & Zero-Shot, One-Shot\\
                Gemini 1.5 Pro & Unknown & 1 Million & Yes & No & Zero-Shot, One-Shot \\ 
                \bottomrule
                \end{tabular}
                \begin{tablenotes}
                    \footnotesize
                    \item [1] Extensible via RoPE embeddings.
                \end{tablenotes}
            \end{threeparttable}
        \end{table*}

    \subsection{Dataset Pre-Processing and Characterization}

        To conduct our experiments, we needed a large body of clinical data that included discharge summaries. For this, we opted for MIMIC-III~\cite{johnsonMIMICIIIFreelyAccessible2016}, a large, publicly available database of de-identified health data from over 40,000 patients gathered from 2001 to 2012 from critical care units of the Beth Israel Deaconess Medical Center. For our work, the standout feature of this database is the addition of clinical notes taken during a patient's stay, including the discharge summaries written upon discharge, that act as a ground truth in our experiments, or, as \citet{adams-etal-2021-whats} suggested, as a \textit{silver truth}.

        During our experimentation with proprietary LLMs such as GPT-3.5, GPT-4, and Gemini 1.5 Pro, we ensured compliance with both the terms of the MIMIC-III Data Use Agreement~\footnote{\href{https://physionet.org/content/mimiciii/view-dua/1.4/}{https://physionet.org/content/mimiciii/view-dua/1.4/}} and applicable European ethical frameworks on the use of AI and Gen AI in research, such as the European Code of Conduct for Research Integrity~\cite{allea-alleuropeanacademiesEuropeanCodeConduct2023} and ERA Guidelines on responsible use of Gen AI in research~\cite{GuidelinesResponsibleUse}. Specifically, we selected API providers that assure that input data is not used for purposes (e.g., models' training) other than the intended computation. These practices align with the guidelines outlined in the article ``Responsible use of MIMIC data with online services like GPT''~\footnote{\href{https://physionet.org/news/post/gpt-responsible-use}{https://physionet.org/news/post/gpt-responsible-use}}.

\begin{figure}[ht]
            \centering
            \includegraphics[width=0.75\linewidth]{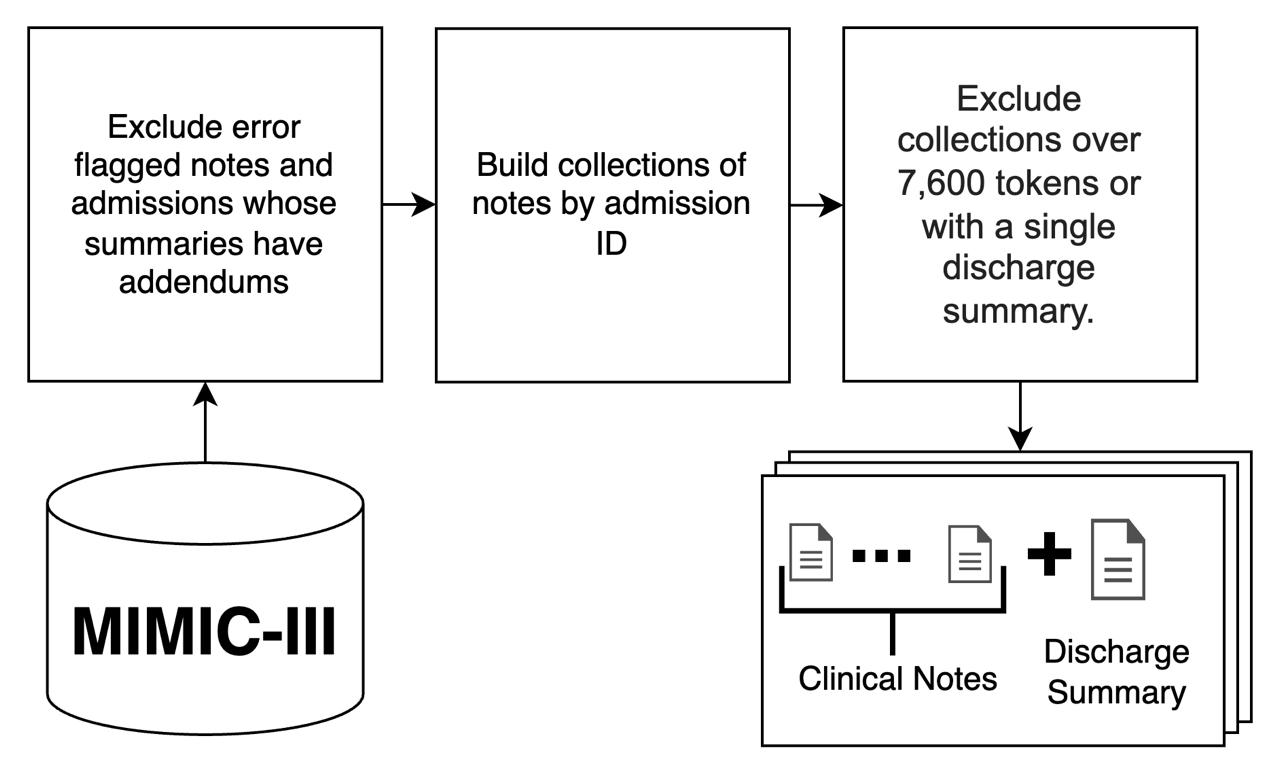}
            \caption{Data Filtering pipeline used to construct our dataset}
            \label{fig:data-filtering}
        \end{figure}

        Although we aimed to use the entire MIMIC-III dataset, we found noise in the data that we needed to filter before proceeding. In this process, shown in Figure~\ref{fig:data-filtering}, we started by removing all the notes flagged as errors by physicians. Next, all the discharge summaries that included additions to the original version, stored as addendums, were also removed to cut down on the larger summaries. This process left us with 1,527,197 notes from 47,006 unique admissions.

        Afterward, we considered the context length of the models and ensured that the data did not surpass the smallest context window of 8,192 tokens. Since both Llama and Mistral use tokenizers based on Sentencepiece~\cite{kudoSentencePieceSimpleLanguage2018a}, a Byte-Pair Encoding (BPE) tokenizer, we used it to filter out collections of notes related to an admission that exceeded 7,600 tokens. This value was chosen for added assurance and to accommodate hardware restrictions since, at this size, admissions could still cause Out-of-Memory errors, impeding our evaluation process. This left us with 21,339 total admissions. These were then randomly split into a train and test set, with 982 admissions reserved as the test set and the remaining 20,357 as training data.

\begin{figure}[ht]
            \centering
            \includegraphics[width=0.8\linewidth]{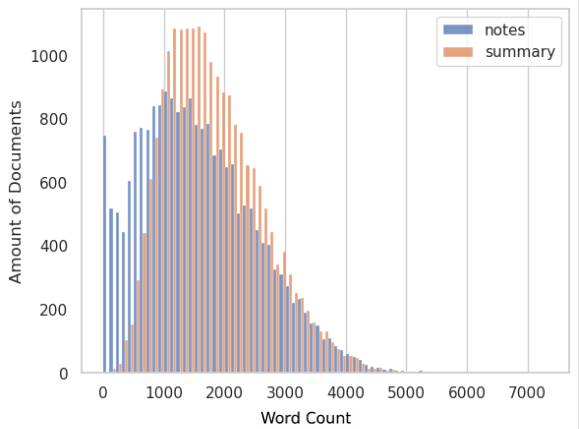}
            \caption{Histogram with word count per document}
            \label{fig:word-count-notes}
        \end{figure}

        We now examine some general features of our dataset. Figure~\ref{fig:word-count-notes} shows a histogram of the word count in both the collection of notes and the summary. On average, a collection of notes on the training set has 1,553 words; on the test set, this number is 2,078. A summary on the training set has an average of 1,803 words, and that value rises to 2,081 on the test set. This indicates that the summaries tend to be larger than the collection of notes on which they were based, which might suggest that they are much more text-heavy than the notes and possibly less information-dense. In contrast, the notes rely on having a lot of information in a more condensed format. This is to be expected, as the discharge summary draws upon information from all notes and possibly contains information not present anywhere else.
    
\begin{figure*}
            \centering
            \begin{subfigure}{0.45\linewidth}
                \includegraphics[width=\linewidth]{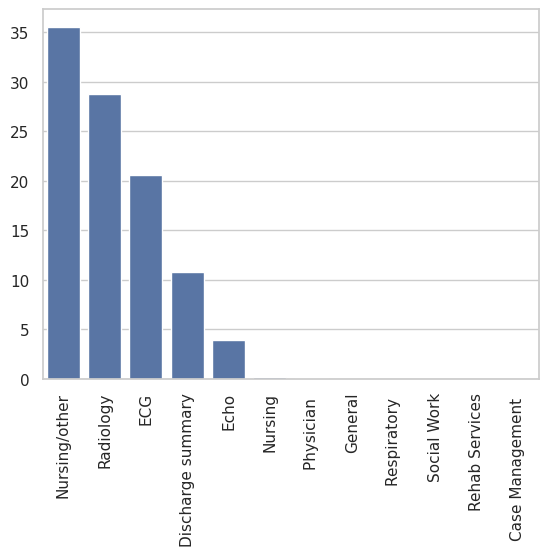}
                \caption{Percentage of notes per category on our dataset}
                \label{fig:count-notes}
            \end{subfigure} \hfill
            \begin{subfigure}{0.45\linewidth}
                \includegraphics[width=\linewidth]{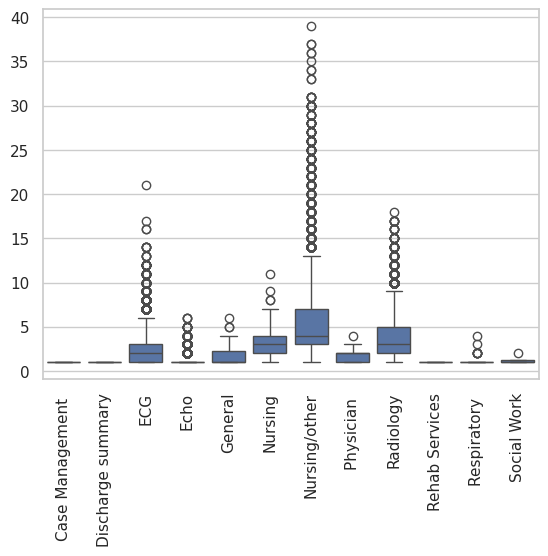}
                \caption{Distribution of notes per Admission on our dataset}
                \label{fig:note-distribution}
            \end{subfigure}
            \begin{subfigure}{0.45\linewidth}
                \includegraphics[width=\linewidth]{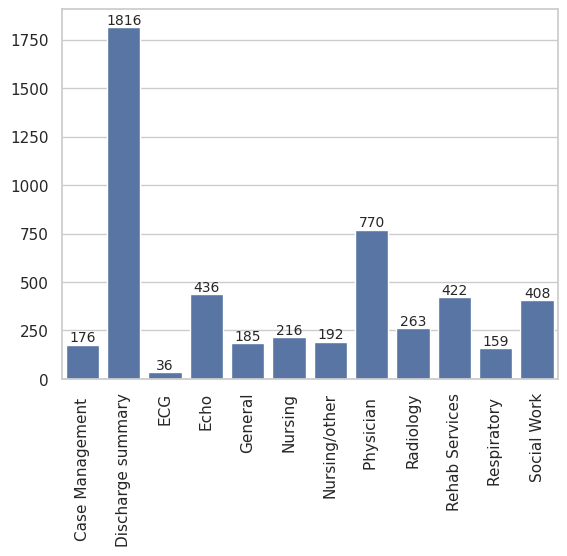}
                \caption{Average amount of words per type of note on our dataset}
                \label{fig:words-type-note}
            \end{subfigure} \hfill
            \begin{subfigure}{0.45\linewidth}
                \includegraphics[width=\linewidth]{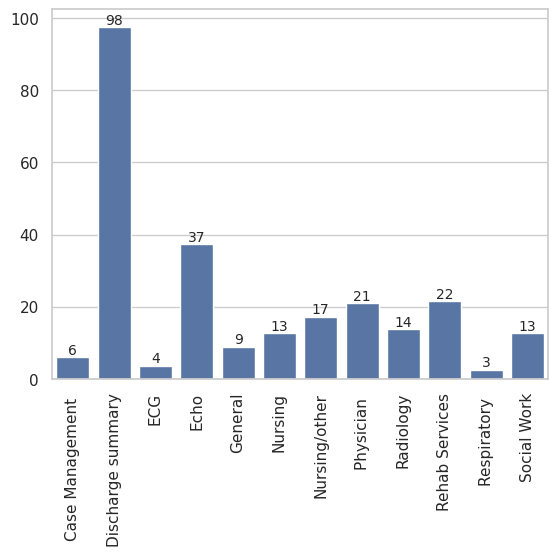}
                \caption{Average amount of sentences per type of note on our dataset}
                \label{fig:sent-type-note}
            \end{subfigure}
        \end{figure*}

        Looking deeper at the collection of notes, we can see their main constituents in Figure~\ref{fig:count-notes}. Nursing, Radiology, and Electrocardiogram (ECG) notes comprise the bulk of our collection. Other categories of notes are much less prevalent, making them less relevant to the overall generation process. Figure~\ref{fig:note-distribution} shows a boxplot with the number of notes per admission.

        If we investigate which documents contain more words or sentences, the trend reverses, with Figure~\ref{fig:words-type-note} showing that the most common documents do not have the most words overall. In fact, the ECG notes contain the least amount of information per note, at just 36 words on average. Nursing and Radiology notes also reverse their trend, and physician notes, much rarer in both datasets, present the most information overall. The amount of sentences, shown in Figure~\ref{fig:sent-type-note}, also corroborates this trend, although nursing notes begin to become more prominent in this matter. The data indicates a trend for many short sentences in these types of notes, while, for example, a Physician's note seems like a more detailed report, with more sentences and words overall.

    \subsection{Experiments Evaluation}

        To assess the performance of each model, we rely on two types of metrics: quantitative metrics from the summarization domain, which will provide a numerical and comparable measure of performance, and a set of qualitative metrics that will offer insights into the relevance, correctness, and reliability of the produced summaries.

        \subsubsection{Quantitative Analysis}
        \label{subsec:metrics}

            To evaluate the quality of the generated summaries, we utilized several well-known metrics. For exact matching, we report BLEU~\cite{papineniBLEUMethodAutomatic2002}, a precision-oriented metric that measures the overlap between a generated summary and the reference text, and ROUGE~\cite{linROUGEPackageAutomatic}, a recall-oriented metric also assessing overlap similarity. We report ROUGE-1 and ROUGE-2, focusing on unigrams and bigrams, and ROUGE-L using the Longest Common Subsequence (LCS)~\cite{linROUGEPackageAutomatic}.

            Since these metrics are limited to exact match scoring, synonyms or slightly different sentences will be disregarded~\cite{kohEmpiricalSurveyLong2022}. To address this, we leveraged BERTScore~\cite{zhangBERTScoreEvaluatingText2019}, which uses BERT embeddings to evaluate semantic similarity, and BLEURT~\cite{sellamBLEURTLearningRobust2020}, which operates similarly but with a model pre-trained on detecting similar sentence structures.

            To analyze sentence repetitions, we used the Type-to-Token Ratio (TTR), defined by $\frac{N_{unique\ words}}{N_{words}}$, which measures vocabulary diversity.

            We also included reference-free metrics, which compare summaries directly with source documents without relying on a ground truth. Specifically, we used BLANC~\cite{vasilyevFillBLANCHumanfree2020}, which measures how well a pre-trained model improves its understanding of source documents after training on the generated summary. This metric correlates with human judgment while bypassing the need for a reference summary.

            For factual consistency, we employed SummaC Conv~\cite{labanSummaCReVisitingNLIbased2022}, a metric that uses a pre-trained Natural Language Inference (NLI) model and a Convolutional Neural Network (CNN) to detect inconsistencies by comparing sentences from the source document and summary.

            Metrics were calculated considering the entire document as a single answer. For proprietary models, a structure to the output was specified, so we also evaluated results by individual sections to evaluate the adherence to the format and identify challenging sections. Such a restriction was not applied to the open-source models due to input size constraints, so a per-section evaluation was not conducted on them. All metrics range from 0 to 1, while BLEURT, according to the implementation description~\cite{bleurt-issue1}, mostly varies from -2 to 1, with some outlier values being possible.

        \subsubsection{Qualitative Analysis}

            To complement the quantitative analysis, we also performed a qualitative evaluation of the results with the help of a clinical expert who read the summaries and evaluated them against three criteria: Completeness, by determining whether the summary captured all relevant information; Correctness, by verifying the factual accuracy of the information presented; and Conciseness, by evaluating whether the summary included any unnecessary information.

            The expert evaluated summaries from three randomly selected admissions. The answers to the questions were given using a 5-point Likert scale, where 1 indicated a ``very bad'' result and 5 a ``very good'' one.

            Additionally, we asked the expert to identify which information was detected as incorrect to possibly determine a pattern of hallucinations in the answers. Moreover, we asked for additional comments the expert deemed necessary.

            Given its small scale, this evaluation should be interpreted as preliminary, and it primarily serves as an initial indication to guide future research directions.
\section{Results and Discussion}
\label{sec:results}

This section examines the main findings of our experiments.

\subsection{Similarity Between Produced and Original Summaries}

    We examine and discuss the performance of all models when evaluating against the gold-standard summaries. Table~\ref{tab:model-scores} presents the scores obtained for each model.
    \begin{table*}[ht]
        \centering
        \footnotesize
        \caption{Scores obtained by each model on the metrics described in Subsection~\ref{subsec:metrics}. Best scores are in bold}
        \label{tab:model-scores}
        \begin{threeparttable}
        \centering
        \begin{tabular}{l c c c c c c c c c}
            \toprule
            \textbf{Model} & \textbf{ROUGE-1} & \textbf{ROUGE-2} & \textbf{ROUGE-L} & \textbf{BLEU} & \textbf{TTR}\tnote{1} & \textbf{BERTScore} & \textbf{BLEURT} & \textbf{BLANC} & \textbf{SummaC} \\ \midrule 
            \textbf{Llama - B}\tnote{2} & 14.16\% & 2.72\% & 8.22\% & 1.30\%* & 21.96\% & 77.83\% & -1.085 & 6.95\% & 53.14\%* \\ 
            \textbf{Llama - F.T}\tnote{3} & 13.99\% & 2.68\% & 8.05\% & 1.23\% & 22.81\% & 77.89\% & -1.084* & 6.67\% & 52.87\% \\ 
            \textbf{Mistral - B} & 21.55\% & 6.14\% & 10.00\% & 0.20\% & 47.93\%* & 79.52\% & -1.085 & \textbf{14.11\%}* & 51.89\% \\ 
            \textbf{Mistral - F.T} & 22.29\%* & 6.34\%* & 10.19\%* & 0.19\% & 47.46\% & 79.66\%* & -1.094 & 13.92\% & 50.24\% \\
            \textbf{GPT-3 - Zero}\tnote{4} & 19.65\% & 7.07\% & 10.63\% & 0.04\% & 51.63\% & 81.87\% & -0.440 & 8.87\% & 50.71\% \\ 
            \textbf{GPT-4 - Zero} & 20.97\% & 7.17\% & 10.89\% & 0.11\% & \textbf{51.77\%}** & 81.75\% & -0.538 & 9.12\% & 51.24\% \\
            \textbf{Gemini - Zero} & 23.33\% & 9.21\% & 12.81\% & 0.32\% & 46.98\% & 82.58\% & -0.469 & 11.69\%** & \textbf{64.94\%}** \\ 
            \textbf{GPT-4 - One}\tnote{4} & 22.30\% & 7.42\% & 11.40\% & 0.33\% & 49.85\% & 82.53\% & -0.484 & 9.90\% & 53.0\% \\ 
            \textbf{Gemini - One} & \textbf{30.90\%}** & \textbf{10.97\%}** & \textbf{15.36\%}** & \textbf{3.07\%}** & 47.14\% & \textbf{85.09\%}** & \textbf{0.0847}** & 10.41\% & 57.07\% \\ 
            \bottomrule
        \end{tabular}
        \begin{tablenotes}
                    \footnotesize
                    \item A single asterisk (*) denotes the best score on an open-source model, whilst a double asterisk (**) denotes the best score on a proprietary model.
                    \item [1] TTR = Type to Token Ratio
                    \item [2] B = Baseline
                    \item [3] F.T = Fine-Tuned
                    \item [4] Zero = Zero-Shot
                    \item [5] One = One-Shot
                \end{tablenotes}
        \end{threeparttable}
    \end{table*}

    \subsubsection{Open-Source Models}

        We begin by analyzing open-source models, starting with the baseline versions and then moving to the fine-tuned ones, assessing performance improvements.

        A first and very noticeable observation, applicable to all models, is that the exact-match scores are low, with the highest barely surpassing 30\%. This is, however, common in similar studies~\cite{palSummarizationGenerationDischarge2022,alaeiAutomatedDischargeSummary2023} as large, multi-document summarization remains a challenge even for state-of-the-art models~\cite{maMultidocumentSummarizationDeep2023}.

        Mistral outperforms Llama on most metrics, only trailing behind in BLEU and SummaC and achieving the same score on BLEURT. Despite underperforming on several other metrics, Llama's baseline model achieved the best BLEU score among the open-source models. However, Llama's vocabulary appears less diverse, as evidenced by the poor TTR score. This is a possible indicative of repeated sentences, a problem common in abstractive summarization~\cite{searleDischargeSummaryHospital2023}. Overall, Mistral takes a significant advantage over Llama, as its results are similar even to those of state-of-the-art proprietary models.

        SummaC results indicate Llama is marginally more factually consistent, with differences of 1.25\% (baseline) and 2.63\% (fine-tuned) over Mistral. However, as SummaC evaluates entailment rather than strict accuracy, its effectiveness diminishes for longer documents~\cite{kohEmpiricalSurveyLong2022}, marking the need for deeper qualitative analysis.

        Mistral showed minor improvement post-fine-tuning, while Llama's performance declined on most metrics. The overall impact of fine-tuning was negligible; Mistral-7B saw only a 0.74\% gain in ROUGE-1 and a slight 0.01\% drop in BLEU. For Llama, ROUGE-1 dropped by 0.17\%, and BLEU by 0.07\%. Changes to the other metrics were also negligible and did not contribute to any significant alteration in performance.
        
        Fine-tuning did not yield significant improvements, likely due to two key factors. First, the 4,096-token limit on generation constrained the output, potentially reducing information richness and structure. However, with an average discharge summary length of 2,081 words, this token limit should accommodate most cases. Second, the lack of a structured output prompt, which aimed to reduce input size due to hardware limitations, may have hindered the models' ability to generate comprehensive outputs.

        Overall, fine-tuning provided little to no performance gains, emphasizing the need to address input and output constraints for better results.

    \subsubsection{Proprietary Models}

        Moving now to the proprietary models, we start our analysis with the results of zero-shot generation, and then we check if moving to a one-shot scenario aids the models.

        Table~\ref{tab:model-scores} reveals a trend in model performance. GPT-3.5, with 175B parameters~\cite{brownLanguageModelsAre2020}, performs slightly worse than the baseline Mistral model, which has only 7B parameters. GPT-4 outperforms GPT-3.5 but is on par with fine-tuned Mistral, with notable differences in TTR and BLEURT.

        Gemini, however, emerges as the top performer in zero-shot, establishing high scores in most categories, particularly in exact-match metrics. Surprisingly, no model surpasses the BLEU score of the baseline Llama, which is more than four times higher than Gemini's.

        Despite improvements, BLANC scores remain largely unchanged for the heavier models. Mistral remains the most consistent here, with proprietary models lagging behind open-source alternatives. In SummaC, only Gemini shows significant improvement in zero-shot, correlating with its strong BLANC performance. Both metrics show similar trends across models, improving or decreasing together.

        When moving to one-shot results, performance notably improves. GPT-4 surpasses previous scores except for TTR, marginally outperforming fine-tuned Mistral. Gemini, however, shows a substantial performance jump, achieving the best overall score, topping six of the eight benchmarks, and nearly tripling Llama's BLEU score. It's also the only model with a positive BLEURT score, making it the top performer despite having subpar results in most areas.

        The subpar exact-match performance can possibly be attributed to the structured format in which they were forced to output. While this may prompt models to include overlooked information, the order of sections may not align with most summaries, affecting exact-match scores. However, relevant information might still be present, as supported by BERTScore, which will be further explored in Section~\ref{sec:qualitative-analysis}.

\subsection{Per-Section Similarity Between Produced and Original Summaries}

    We now look deeper into the proprietary models and consider each section separately. Since each proprietary model was asked to output in a structured format, it is reasonable to assume the answer was generated with this structure, a hypothesis we corroborated when analyzing the results with regular expressions. If a section was missing in the original or generated summary, an empty string was used for comparison. For BLANC and SummaC, we used the entire source document as a reference to evaluate the impact of individual sections on model performance.

    Table~\ref{tab:header-scores} presents the analysis, showing the best scores per section in bold. Gemini, using one-shot prompting, emerged as the top performer with 73 best results (including ties), followed by GPT-3 with 53 best results. The other models showed more consistent but lower performance. Notably, while Gemini excelled in exact-match metrics (ROUGE, BLEU), GPT-3 performed better in TTR, BERTScore, BLANC, and SummaC. Gemini's strong exact-match scores suggest it closely mimics discharge summary language, whereas GPT-3 scores better in vocabulary diversity.

    {\setlength\rotFPtop{0pt plus 1fil}
    \begin{sidewaystable*}
        \tiny
        \centering
        \caption{Scores (ROUGE-\{1/2/L\}/BLEU/TTR/BertScore/BLEURT/BLANC/SummaC) obtained by proprietary models under each section from the zero-shot prompt. Best scores per section and metric are in bold}
        \label{tab:header-scores}
        \begin{tabular}{p{0.12\linewidth} p{0.145\linewidth} p{0.145\linewidth} p{0.145\linewidth} p{0.145\linewidth} p{0.145\linewidth}}
        \toprule
            \textbf{Section} & \textbf{GPT-3 - Zero-Shot} & \textbf{GPT-4 - Zero-Shot} & \textbf{Gemini - Zero-Shot} & \textbf{GPT-4 - One-Shot} & \textbf{Gemini - One-Shot} \\ \midrule
            \textbf{Service} & 7.42\%, 0.42\%, 7.35\%, \textbf{0.18\%}, 99.45\%, 80.10\%, -0.470, -0.27\%, 56.87\% & 15.55\%, \textbf{1.26\%}, 15.49\%, 0.10\%, 98.05\%, 80.10\%, -0.295, -0.31\%, \textbf{58.99\%} & 11.04\%, 0.51\%, 10.95\%, 0\%, 95.79\%, 78.87\%, -0.510, \textbf{-0.05\%}, 50.85\% & 28.29\%, 0.66\%, 28.11\%, 0\%, \textbf{99.71\%}, 83.11\%, -0.026, -0.28\%, 46.07\% & \textbf{43.37\%}, 0.10\%, \textbf{43.38\%}, 0\%, 99.60\%, \textbf{87.93\%}, \textbf{0.250}, -0.32\%, 38.24\% \\ \midrule
            \textbf{Chief Complaint} & 12.75\%, 5.54\%, 12.24\%, 1.08\%, \textbf{96.33\%}, 60.57\%, -1.021, \textbf{0.25\%}, \textbf{70.69\%} & \textbf{14.71\%}, 6.53\%, 14.16\%, 1.01\%, \textbf{96.33\%}, 60.73\%, -0.951, 0.05\%, 70.57\% & 14.12\%, 6.68\%, 13.69\%, 1.08\%, 90.47\%, 54.75\%, -0.920, -0.05\%, 69.94\% & 14.33\%, 6.54\%, 13.87\%, 0.93\%, 96.02\%, \textbf{60.79\%}, -0.966, -0.04\%, 70.16\% & 14.57\%, \textbf{7.74\%}, \textbf{14.40\%}, \textbf{1.96\%}, 94.10\%, 59.72\%, \textbf{-0.788}, -0.25\%, 61.89\% \\ \midrule
            \textbf{Major Surgical or Invasive Procedure} & 20.77\%, 2.90\%, 20.01\%, 1.10\%, \textbf{96.72\%}, 61.25\%, -0.896, -0.04\%, 60.65\% & 20.73\%, 4.27\%, 19.53\%, \textbf{2.53\%}, 94.75\%, 60.45\%, -0.970, \textbf{0.03\%}, \textbf{63.13\%} & 13.26\%, 3.27\%, 12.71\%, 2.02\%, 88.83\%, 53.84\%, -0.978, \textbf{0.03\%}, 59.75\% & 23.23\%, \textbf{4.39\%}, 22.41\%, 1.83\%, 95.28\%, 61.48\%, -0.878, -0.07\%, 60.06\% & \textbf{25.02\%}, 3.49\%, \textbf{24.60\%}, 0.90\%, 96.62\%, \textbf{61.64\%}, \textbf{-0.705}, -0.19\%, 52.09\% \\ \midrule
            \textbf{History of Present Illness} & 12.52\%, 3.08\%, 8.83\%, 0.41\%, 81.78\%, \textbf{58.93\%}, \textbf{-1.028}, \textbf{3.03\%}, \textbf{52.61\%} & 10.93\%, 2.88\%, 7.98\%, 0.11\%, \textbf{83.82\%}, 56.72\%, -1.060, 1.04\%, 45.10\% & \textbf{13.52\%}, \textbf{4.09\%}, \textbf{9.50\%}, 1.27\%, 74.27\%, 53.84\%, -1.172, 1.94\%, 49.60\% & 9.22\%, 2.58\%, 6.98\%, 0.06\%, 69.02\%, 47.14\%, -1.085, 1.06\%, 41.53\% & 9.86\%, 2.44\%, 6.40\%, \textbf{1.45\%}, 42.74\%, 36.30\%, -1.098, 1.03\%, 35.40\% \\ \midrule
            \textbf{Past Medical History} & 6.27\%, 1.96\%. 5.52\%, 0.02\%, \textbf{92.23\%}, \textbf{57.84\%}, -1.312, \textbf{-0.04\%}, \textbf{60.12\%} & 6.43\%, 1.98\%, 5.47\%, 0.08\%, 86.91\%, 55.33\%, -1.248, -0.24\%, 53.01\% & \textbf{6.94\%}, \textbf{2.28\%}, \textbf{6.09\%}, \textbf{0.25\%}, 84.26\%, 53.72\%, -1.294, -0.10\%, 55.09\% & 5.19\%, 1.70\%, 4.38\%, 0.05\%, 71.91\%, 46.43\%, -1.222, -0.15\%, 46.98\% & 3.84\%, 1.07\%, 3.27\%, 0.08\%, 39.99\%, 26.25\%, \textbf{-1.118}, -0.24\%, 33.86\% \\ \midrule
            \textbf{Allergies} & 2.63\%, 0.57\%, 2.48\%, 0\%, \textbf{98.94\%}, 58.13\%, \textbf{-0.961}, \textbf{-0.23\%}, \textbf{75.86\%} & \textbf{12.56\%}, 5.73\%, \textbf{12.37\%}, 0\%, 97.56\%, \textbf{58.94\%}, -1.132, -0.41\%, 65.96\% & 3.56\%, 0.89\%, 3.39\%, 0.64\%, 91.64\%, 51.74\%, -1.049, -0.36\%, 55.53\% & 9.60\%, 5.11\%, 9.47\%, 0.82\%, 97.62\%, 58.57\%, -1.102, -0.42\%, 61.74\% & 12.27\%, \textbf{7.02\%}, 11.91\%, \textbf{5.27\%}, 97.76\%, 58.46\%, -1.144, -0.55\%, 48.23\% \\ \midrule
            \textbf{Medications on Admission} & 1.32\%, 0.06\%, 1.15\%, 0\%, 92.28\%, 54.15\%, -1.500, \textbf{-0.03\%}, \textbf{68.49\%} & 1.73\%, 0.10\%, 1.51\%, 0\%, 92.25\%, 54.07\%, -1.488, -0.14\%, 63.08\% & 1.18\%, 0.06\%, 1.00\%, 0\%, 89.66\%, 48.53\%, -1.407, -0.38\%, 54.67\% & 1.50\%, 0.07\%, 1.28\%, 0\%, \textbf{93.53\%},\textbf{ 54.28\%}, -1.459, -0.13\%, 62.45\% & \textbf{4.94\%}, \textbf{0.78\%}, \textbf{4.05\%}, \textbf{0.30\%}, 62.06\%, 43.00\%, \textbf{-1.21}, -0.44\%, 38.26\% \\ \midrule
            \textbf{Family History} & 0.45\%, 0.05\%, 0.45\%, 0\%, \textbf{99.31\%}, \textbf{57.79\%}, \textbf{-0.895}, -0.17\%, \textbf{77.54\%} & 1.47\%, 0.29\%, 1.39, 0\%, 98.16\%, 56.97\%, -1.076, \textbf{-0.14\%}, 66.37\% & 1.13\%, 0.30\%, 1.06\%, 0\%, 91.48\%, 50.98\%, -0.984, -0.30\%, 55.83\% & 0.81\%, 0.06\%, 0.77\%, 0\%, 97.22\%, 56.75\%, -1.006, -0.24\%, 66.34\% & \textbf{4.62\%}, \textbf{0.96\%}, \textbf{3.99\%}, \textbf{0.53\%}, 86.66\%, 54.47\%, -1.204, -0.26\%, 48.64\% \\ \midrule
            \textbf{Social History} & 3.06\%, 0.51\%, 2.74\%, 0.05\%, 96.72\%, 57.53\%, -1.367, \textbf{0.12\%}, \textbf{65.21\%} & 3.53\%, 0.63\%, 3.09\%, 0.09\%, 96.69\%, 57.31\%, -1.400, -0.04\%, 60.66\% & 3.85\%, 0.87\%, 3.35\%, 0.31\%, 88.54\%, 52.06\%, -1.395, -0.10\%, 53.08\% & 2.85\%, 0.53\%, 2.44\%, 0.07\%, \textbf{97.11\%}, 57.42\%, -1.380, -0.04\%, 62.60\% & \textbf{8.99\%}, \textbf{1.75\%}, \textbf{7.24\%}, \textbf{1.42\%}, 91.17\%, \textbf{58.13\%}, \textbf{-1.262}, -0.03\%, 47.76\% \\ \midrule
            \textbf{Physical Exam} & 3.61\%, 0.26\%, 2.77\%, 0.08\%, 84.53\%, \textbf{57.10\%}, \textbf{-1.294}, \textbf{1.68\%}, 54.37\% & 3.59\%, 0.31\%, 2.68\%, 0.09\%, \textbf{86.47\%}, 56.65\%, -1.371, 0.64\%, 52.53\% & 3.51\%, 0.32\%, 2.35\%, 0.26\%, 78.29\%, 50.88\%, -1.480, 1.10\%, \textbf{67.42\%} & 3.93\%, 0.31\%, 2.77\%, 0.19\%, 83.59\%, 56.72\%, -1.398, 1.05\%, 55.02\% & \textbf{8.33\%}, \textbf{1.36\%}, \textbf{5.85\%}, \textbf{2.19\%}, 75.84\%, 56.82\%, -1.308, 0.7\%, 66.33\% \\ \midrule
            \textbf{Pertinent Results} & 2.31\%, 0.77\%, 1.89\%, 0\%, \textbf{88.16\%}, 54.47\%, -1.433, 1.99\%, 46.20\% & 3.57\%, 1.20\%, 2.59\%, 0.14\%, 81.83\%, 54.28\%, -1.415, 3.86\%, 47.99\% & 5.10\%, 2.50\%, 4.11\%, 2.19\%, 68.24\%, 49.71\%, -1.410, \textbf{5.66\%}, 65.51\% & 3.98\%, 1.31\%, 2.86\%, 0.34\%, 78.85\%, 54.50\%, -1.409, 4.72\%, 47.75\% & \textbf{7.79\%}, \textbf{2.74\%}, \textbf{5.57\%}, \textbf{3.17\%}, 74.46\%, \textbf{54.53\%}, \textbf{-1.329}, 3.96\%, \textbf{67.96\%} \\ \midrule 
            \textbf{Brief Hospital Course} & 8.39\%, 1.15\%, 5.86\%, 0.03\%, 82.32\%, 58.64\%, \textbf{-1.023}, 1.42\%, \textbf{51.84\%} & 8.49\%, 1.06\%, 5.69\%, 0.04\%, \textbf{82.68\%}, 57.99\%, -1.074, 1.08\%, 47.97\% & 11.49\%. 2.84\%, 7.59\%, 0.31\%, 71.52\%, 53.91\%, -1.154, 1.70\%, 46.72\% & 10.07\%, 1.29\%, 6.39\%, 0.16\%, 77.87\%, 57.87\%, -1.033, 1.77\%, 48.74\% & \textbf{14.57\%}, \textbf{3.45\%}, \textbf{9.10\%}, \textbf{1.87\%}, 69.13\%, \textbf{58.69\%}, -1.084, \textbf{1.90\%}, 43.83\% \\ \midrule
            \textbf{Discharge Diagnosis} & 10.11\%, 3.38\%, 9.11\%, 0.65\%, \textbf{93.11\%}, \textbf{59.48\%}, -1.124, \textbf{1.08\%}, 57.62\% & 12.78\%, 4.75\%, 11.32\%, 0.98\%, 90.52\%, 59.42\%, -1.110, -0.45\%, 56.10\% & 9.16\%, 4.05\%, 8.50\%, 0.87\%, 87.75\%, 53.07\%, -1.204, -0.27\%, 63.49\% & 13.02\%, 4.85\%, 11.47\%, 1.13\%, 86.17\%, 58.98\%, \textbf{-1.079}, 0.34\%, 49.97\% & \textbf{13.52\%}, \textbf{5.42\%}, \textbf{12.19\%}, \textbf{1.86\%}, 88.78\%, 59.24\%, -1.094, -0.43\%, \textbf{69.47\%} \\ \midrule
            \textbf{Discharge Medications} & 0.75\%, 0.23\%, 0.73\%, \textbf{0\%}, \textbf{96.52\%}, 52.60\%, -1.428, -0.11\%, \textbf{70.86\%} & 1.00\%, 0.12\%, 0.89\%, \textbf{0\%}, 92.14\%, \textbf{52.81\%}, -1.529, -0.18\%, 60.77\% & 0.56\%, 0.08\%, 0.49\%, \textbf{0\%}, 90.65\%, 47.54\%, -1.344, -0.33, 54.39\% & 0.81\%, 0.11\%, 0.71\%, \textbf{0\%}, 92.43\%, 52.63\%, -1.474, \textbf{-0.08\%}, 58.67\% & \textbf{3.59\%}, \textbf{0.84\%}, \textbf{2.83\%}, \textbf{0\%}, 47.73\%, 34.15\%, \textbf{-1.028}, -0.39\%, 32.51\% \\ \midrule
            \textbf{Discharge Disposition} & 8.06\%, 0.04\%, 8.09\%, \textbf{0\%}, \textbf{98.97\%}, \textbf{60.04\%}, -1.385, -0.18\%, \textbf{66.10\%} & 7.45\%, 0.68\%, 7.45\%, \textbf{0\%}, 97.88\%, 59.30\%, -1.195, -0.42\%\%, 62.86\% & 8.63\%, 0.65\%, 8.62\%, \textbf{0\%}, 91.14\%, 53.99\%, -0.835, -0.38\%, 49.68\% & 10.72\%, 0.74\%, 10.67\%, \textbf{0\%}, 97.03\%, 59.77\%, -1.110, \textbf{-0.13\%}, 52.92\% & \textbf{20.71\%}, \textbf{4.80\%}, \textbf{20.72\%}, \textbf{0\%}, 71.07\%, 47.56\%, \textbf{-0.389}, \textbf{-0.13\%}, 45.94\% \\ \midrule
            \textbf{Discharge Instructions} & 3.82\%, 0.76\%, 3.12\%, 0.01\%, \textbf{94.90\%}, \textbf{57.38\%}, -1.278, 0.15\%, 58.71\% & 3.99\%, 0.44\%, 3.13\%, 0.08\%, 92.07\%, 56.82\%, -1.210, -0.22\%, \textbf{61.58\%} & 1.12\%, 0.06\%, 0.99\%, 0\%, 91.14\%, 50.64\%, -1.323, -0.36\%, 54.78\% & 5.63\%, 0.74\%, 3.97\%, 0.39\%, 85.71\%, 57.22\%, \textbf{-1.191}, -0.05\%, 55.66\% & \textbf{6.53\%}, \textbf{1.37\%}, \textbf{4.56\%}, \textbf{2.18\%}, 81.03\%, 55.79\%, -1.295, \textbf{0.19\%}, 43.68\% \\ \midrule
            \textbf{Discharge Condition} & \textbf{10.12\%}, 0.08\%, \textbf{10.13\%}, \textbf{0\%}, \textbf{99.07\%}, \textbf{59.79\%}, \textbf{-0.868}, -0.43\%, 34.36\% & 4.39\%, \textbf{0.12\%}, 4.06\%, \textbf{0\%}, 96.24\%, 58.98\%, -1.440, \textbf{0.12\%}, \textbf{62.69\%} & 6.24\%, 0\%, 6.17\%, \textbf{0\%}, 91.05\%, 52.80\%, -0.956, -0.48\%, 43.47\% & 4.67\%, 0.08\%, 4.40\%, \textbf{0\%}, 94.11\%, 58.91\%, -1.449, -0.01\%, 62.05\% & 9.77\%, 0.07\%, 9.75\%, \textbf{0\%}, 85.50\%, 52.19\%, -0.908, -0.52\%, 38.53\% \\ \midrule
            \textbf{Follow-up Instructions} & 4.06\%, 1.20\%, 3.70\%, 0\%, \textbf{97.76\%}, \textbf{55.58\%}, -1.409, \textbf{0.18\%}, \textbf{54.36\%} & 4.62\%, \textbf{1.56\%}, 4.07\%, 0.05\%, 93.07\%, 55.44\%, -1.392, 0.10\%, 50.09\% & 1.42\%, 0.49\%, 1.30\%, 0\%, 90.79\%, 49.12\%, -1.396, -0.23\%, \textbf{54.36\%} & \textbf{4.79\%}, 1.44\%, \textbf{4.17\%}, 0.06\%, 91.47\%, 55.33\%, -1.390, 0.12\%, 47.91\% & 3.98\%, 1.45\%, 3.29\%, \textbf{0.40\%}, 41.92\%, 29.91\%, \textbf{-1.072}, -0.06\%, 30.85\% \\
            \bottomrule
    \end{tabular}
    \end{sidewaystable*}}

    In soft-overlap metrics, BLEURT and BERTScore, both models were competitive, but BERTScore favored GPT-3, while BLEURT gave a slight edge to Gemini. BLANC corroborated this trend, with GPT-3 scoring highest in reference-free evaluations, suggesting it enhances understanding of the original text. However, many BLANC scores were negative, indicating some sections degraded understanding. SummaC also favored GPT-3, with 11 of the 18 best scores, indicating its consistency across sections.

    In comparison with the ``Discharge Me!'' task from Xu et al.~\cite{xu-etal-2024-overview}, which focused on the ``Brief Hospital Course'' and ``Discharge Instructions'' sections, we computed the average scores between the relevant sections for our models. The top performer remains Gemini with One-Shot prompts, which showed subpar results on the exact-match metrics, with 10.55\% ROUGE-1, 2.41\% ROUGE-2, 6.83\% ROUGE-L, and 2.025\% BLEU. Such scores would place it among the last positions on the leaderboard. However, our BERTScore was better than the top performer of the challenge, with an improvement of 13.44\%. This indicates that Gemini is providing correct output, with a meaning similar to the ground-truth, but will tend to use starkly different wording compared to what is expected.
    
    Considering section complexity, sections like ``Service,'' ``Major Surgical or Invasive Procedure'', and ``Discharge Disposition'' yielded good scores, while categories like ``Discharge Medications'', ``Follow-up Instructions'', and ``Past Medical History'' were more challenging. This aligns with expectations, as these sections often include information omitted during the hospital stay, with discharge details typically written at the time of discharge rather than during the hospital course.
    
\subsection{Similarity Between Produced Summaries and Clinical Notes}

    The lower scores of Tables~\ref{tab:model-scores} and~\ref{tab:header-scores} prompted us to examine the degree of overlap between our produced summaries and the original clinical notes instead of comparing them against the gold-standard discharge summaries. The results are presented in Table~\ref{tab:overlap-scores}. Since we measured only the overlap, we omitted BLANC, SummaC, and TTR, all reference-free metrics, from this evaluation.

    \begin{table*}[ht]
        \centering
        \caption{Overlap metrics of each model against the original notes. Best scores are in bold}
        \label{tab:overlap-scores}
        \begin{tabular}{l c c c c c c c c c}
            \toprule
            \textbf{Model}  & \textbf{ROUGE-1} & \textbf{ROUGE-2} & \textbf{ROUGE-L} & \textbf{BLEU} & \textbf{BERTScore} & \textbf{BLEURT} \\ \midrule
            \textbf{Llama - B} & 19.26\% & 6.29\% & 11.86\% & 1.88\% & 78.41\% & -0.904 \\
            \textbf{Llama - F.T} & 19.77\% & 6.62\% & 12.09\% & \textbf{1.99\%} & 78.44\% & -0.900 \\
            \textbf{Mistral - B} & \textbf{31.19\%} & \textbf{15.87\%} & \textbf{19.02\%} &  0.1\% & 80.12\% & -0.927 \\
            \textbf{Mistral - F.T} & 31.13\% & 15.49\% & 18.57\% & 0.09\% & 80.14\% & -0.937  \\
            \textbf{GPT-3 - Zero} & 20.36\% & 8.07\% & 10.60\% & 0\% & 79.89\% & \textbf{-0.776} \\
            \textbf{GPT-4 - Zero} & 21.55\% & 7.80\% & 10.49\% & 0.02\% & 79.95\% & -0.829 \\
            \textbf{Gemini - Zero} & 24.87\% & 13.41\% & 14.59\% & 0.09\% & 80.82\% & -0.824 \\
            \textbf{GPT-4 - One} & 23.68\% & 8.67\% & 11.35\% & 0.06\% & 80.92\% & -0.837 \\
            \textbf{Gemini - One} & 26.31\% & 11.99\% & 13.93\% & 1.38\% & \textbf{81.78\%} & -0.808 \\
            \bottomrule
        \end{tabular}
    \end{table*}

    Overlap is more prominent in the open-source models, with Mistral having the highest scores overall and considerable results in the exact-match metrics. BLEU also is comparatively high for Llama, with Mistral falling short in this score. This trend in performance leads us to believe that, despite the fine-tuning process not bringing any significant advantage to generate a discharge summary, it led to them modeling the text better and becoming more efficient at transcribing features from the text. Despite these improvements in extracting information from the text, their performance was worse when reasoning over the information and synthesizing it correctly into a discharge summary, as evidenced by the contrast in scores.

    Surprisingly, the proprietary models show little improvement compared to the open-source ones. In the exact-match metrics, almost all results from the GPT models are only slightly above Llama's. One hypothesis that might explain this is that, because of their size and reasoning ability, these models could distill the data from the notes into a discharge summary, inferring from it and giving results akin to an actual document. The structured output could have also helped in this process by guiding the output and forcing the models to adapt to a defined structure. 

    Considering the soft-overlap metrics of BERTScore and BLEURT, the evaluations show a stark contrast. Although BERTScore remained stable for all models, BLEURT gives lower scores in the proprietary models. In Table~\ref{tab:model-scores}, an improvement is seen in BLEURT when moving to the proprietary variants, with Gemini even reaching a positive result. However, in Table~\ref{tab:overlap-scores}, the scores don't vary substantially and are consistent throughout all models.
    
    Overall, the results indicate that the open-source models tend to better transcribe information directly from the source text into the discharge summary, whilst the proprietary models tend to generate sentences that convey the same meaning as the original notes but use different terms.

\subsection{Completeness, Correctness, and Conciseness of the Produced Summaries}
\label{sec:qualitative-analysis}

    We now look at the results of the qualitative evaluation conducted by our medical expert. We used a 5-point Likert scale, where the worst score was a 1, and the best possible score was a 5. Table~\ref{tab:qualitative-scores} presents the scores of each criterion for each document.

     \begin{table}[ht]
         \centering
         \caption{Results of the qualitative evaluation per document. Best scores are in bold}
         \label{tab:qualitative-scores}
         \begin{tabular}{l c c c}
             \toprule
             \textbf{Model} & \textbf{Completeness} & \textbf{Correctness} & \textbf{Conciseness} \\ \midrule
             Llama - Baseline & \{1, 1, 1\} & \{1, 1, 1\} & \{1, 1, 1\} \\
             Llama - Finetuned & \{1, 1, 1\} & \{1, 1, 1\}& \{1, 1, 1\} \\
             Mistral - Baseline & \{2, 1, 1\} & \{3, 1, 1\} & \{2, 1, 1\} \\
             Mistral - Finetuned & \{2, 2, 2\} & \{3, 2, 2\} & \{2, 2, 2\} \\
             GPT3 - Zero-Shot  & \{2, 2, 2\} & \{2, 1, 2\} & \{2, 2, 2\} \\
             GPT4 - Zero-Shot & \{2, 2, 2\} & \{2, 2, 2\} & \{2, 2, 2\} \\
             Gemini - Zero-Shot & \{2, 2, 2\} & \{2, 2, 1\} & \{2, 2, 2\} \\
             GPT-4 - One-Shot & \textbf{\{3, 3, 3\}} & \textbf{\{3, 3, 3\}} & \textbf{\{3, 3, 3\}} \\
             Gemini - One-Shot & \{3, 2, 3\} & \{2, 2, 2\} & \{3, 2, 3\} \\
             \bottomrule
         \end{tabular}
     \end{table}

    Across three samples, GPT-4 with one-shot prompting consistently achieved the highest scores across all categories despite overall subpar results. Both Llama versions and baseline Mistral performed poorly, often scoring at the lowest end of the scale. Fine-tuned Mistral showed improvement, achieving scores comparable to zero-shot approaches, but one-shot strategies remained the best performers.

    One major issue noted by the clinician was the lack of completeness in summaries. No model reliably included all information from the gold-standard summaries. Looking deeper into this issue, we see that it is likely a problem with the data itself, as some necessary information is missing from the notes. This notion, tied with the results from Table~\ref{tab:header-scores}, reinforces our belief that information such as what to do upon discharge and any follow-up necessary is only determined when the summary is written and not commonly present in the remaining notes. This statement is further reinforced by the work of \citet{adams-etal-2021-whats}, which mentions that clinicians receive little formal instruction in documenting patient information and are pressed for time when doing so. Nevertheless, based on the expert's notes, GPT-4 and Gemini, with one-shot prompting, did a reasonable job of capturing all available information, even though some critical details might be absent.

    Correctness was another frequent issue. Llama often produced repetitive, incoherent, or irrelevant outputs despite reasonable BLEU and BERTScore metrics, likely due to its use of medical terminology in the produced sentences. Mistral showed a slight improvement but missed key details, though it reliably extracted relevant information from notes. However, it sometimes deviated from standard summary formats, presenting outputs as letters to physicians.

    Conciseness appears to follow the trend of completeness. Proprietary models, particularly in one-shot scenarios, showed substantial improvements in correctness. However, they occasionally included hallucinated details not found in the notes, such as a recommendation for a patient to follow a ``low-residue diet'' when no such information was given. Gemini was more prone to such hallucinations, possibly due to data leaking from the example summary into the response, leading to incoherence.

    Overall, this analysis was consistent with our previous results, giving an advantage to the proprietary models that use a one-shot approach. Although our quantitative metrics gave the edge to Gemini, meaning that it is the most similar to the actual summaries, this analysis preferred GPT-4 by some margin, making them the two most viable and most promising approaches to clinicians.

\subsection{Inference Speed Analysis}

    Quick generation is a desirable and important factor when dealing with multiple requests. Therefore, we also evaluated the models' inference speed, relying on the methodology from Deci AI~\cite{deciCorrectWayMeasure2023}. In Table~\ref{tab:inference-times}, we report the results after five warmup steps, using 100 repetitions of an example chosen randomly. For the proprietary models, as these rely on APIs, such an analysis is impossible.

    \begin{table}[ht]
        \centering
        \caption{Inference times for the open-source models}
        \label{tab:inference-times}
        \begin{tabular}{l c c}
            \toprule
            \textbf{Model} & $\overline{t}_{inference}$ & $\sigma_{inference}$ \\ \midrule
            \textbf{Llama-Baseline} & 84,052,84ms & 60,369.35ms \\
            \textbf{Llama-Finetuned} & 71,694.74ms & 59,991.02ms \\
            \textbf{Mistral-Baseline} & 18,389.22ms & 24.07ms \\
            \textbf{Mistral-Finetuned} & 17,732.36ms & 33.46ms \\
            \bottomrule
        \end{tabular}
    \end{table}

    There is a clear advantage in Mistral, which presented the best scores by far, with our fine-tuned version being four times faster on average than Llama. Surprisingly, besides being very slow in its generation, Llama also presents a considerably higher degree of variability in the scores. A probable answer to this is the use of RoPE embeddings~\cite{suRoFormerEnhancedTransformer2023}, which possibly slows down the computations where the length exceeds the model's predefined context window of 4,096 tokens.
    
\section{Considerations for Real-World Deployment}

    Considering the results from the previous sections, we can consider the two most viable approaches, which we deem to be Gemini 1.5 Pro and GPT-4, using a one-shot prompt approach. However, there are some caveats to these options that we can further explore.

    At the time of writing, the prices each model charges are available in Table~\ref{tab:model-pricing}. Gemini 1.5 Pro costs 1.5\$ per million tokens of input and 5\$ per million tokens of output, considering that the total amount of tokens does not exceed 128,000 tokens. The version of GPT-4 that we tested on, GPT-4 Turbo, is even more expensive, at 10\$ per million tokens on input and 30\$ on output. Besides, both options are rate-limited, with Gemini maxing out at 1,000 requests per minute and 4 million tokens per minute. GPT-4, despite allowing 10,000 requests per minute, limits the input to 2 million tokens per minute in its highest tier. These figures would have to be considered carefully in any real-world deployment.

    \begin{table}[ht]
        \centering
        \caption{Pricing and Rate Limits of proprietary models}
        \label{tab:model-pricing}
        \begin{tabular}{l c c c c}
            \toprule
            \textbf{Model} & \textbf{Price / 1M tokens} & \textbf{T.P.M} & \textbf{R.P.M} & \textbf{R.P.D} \\ \midrule
            GPT-3 & 0.5\$ / 1.5\$ & 50M & 10,000 & None \\
            GPT-4 & 10\$ / 30\$ & 2M & 10,000 & None \\
            Gemini 1.5 Pro & 1.5\$ / 5\$ & 4M & 1000 & None \\
            \bottomrule
        \end{tabular}
    \end{table}

    A possible alternative would be to use a self-hosted model like the fine-tuned Mistral, which offers cost benefits by eliminating generation fees but with the tradeoff of reduced performance, more frequent hallucinations, and missing information compared to the proprietary systems. Llama 2, in the tested size, is not recommended due to its poor performance and inefficiency. Another issue with self-hosting is that it requires significant infrastructure to handle long inputs and scale for hospital-wide use, creating significant maintenance costs and speed tradeoffs. For instance, Mistral took over eight hours to generate 982 summaries, while Llama was over four times slower, making both impractical for clinical use without substantial resources. Additionally, rapid innovation in AI means models can become obsolete within a year, limiting long-term viability.

    It is also important to consider privacy and data handling in closed-source models. Clinical notes are classified as sensitive information, protected under the General Data Protection Regulation (GDPR)~\cite{RegulationEU20162016}. Unless these models ensure GDPR compliance, they cannot be used for automatic generation. 

    Nonetheless, although it is possible to use both alternatives in a real-world setting, this problem still needs significant research efforts and advancements to be considered feasible for adoption. Currently, a hybrid approach where AI generates a draft for physicians to review and edit appears as a more reliable option, reducing clinician workload while improving care quality without compromising accuracy.

    Our study has limitations, which we will now highlight. Hardware was a significant hindrance, as it restricted this study regarding the data and model selection. Due to this, many data points were discarded from MIMIC, and model selection and training were also impacted. Additionally, only MIMIC-III data was used, which may not generalize to other hospitals with different summarization styles. In this topic, we also did not consider the inherently subjective nature of importance. What is critical in one care unit may not be relevant in another, and different professionals have unique summarisation styles, which can vary significantly between them.

    Our human analysis was conducted by a medical expert, using just three samples from each model, limiting the generalizability of the results. Additionally, what might be relevant for one clinician can be negligible for another, so having the evaluation done by more people would be an advantage, as the scores would be more reliable. Furthermore, language barriers and unfamiliarity with American hospital abbreviations may have negatively impacted the assessment, as our clinical expert is Portuguese and works within Portugal's health service.

    Finally, it is also important to note that we did not thoroughly engineer our hyperparameters when fine-tuning our models, nor did we thoroughly evaluate settings such as temperature in the proprietary models, which can lead to changes in the generated responses.
\section{Conclusions and Future Work}
\label{sec:conclusion}
    
    This work addressed the challenge of generating discharge summaries from prior clinical notes, aiming to improve the efficiency of patient documentation for clinicians. We evaluated five models, ranging from small open-source versions to large proprietary systems, using a comprehensive suite of metrics alongside analyses of document overlap and inference speed. While proprietary models showed superior performance, certain open-source options, such as the fine-tuned Mistral, demonstrated competitive results. However, the experiments indicate that significant advancements are still needed for real-world applicability.  

    Future research could explore a broader range of models, including other open-source options like Gemma or larger variants such as Llama-3 70B or Llama-3 400B~\cite{llama3modelcard}. Using locally sourced data and experimenting with datasets in different languages, particularly low-resource ones, could enhance adaptability and robustness in diverse clinical settings. On the topic of data, creating synthetic datasets would also enable unrestricted distribution, facilitating research.

    Addressing missing data in clinical notes presents another avenue for improvement. Introducing alternative methods for capturing detailed information, such as voice recordings, could enrich the dataset. With advancements in multimodal data research, exploring new approaches to discharge summary generation, including multimodal systems, holds significant potential~\cite{jangraSurveyMultimodalSummarization2023}.

\bmhead*{Acknowledgements}
This work is co-financed by Component 5 - Capitalization and Business Innovation, integrated in the Resilience Dimension of the Recovery and Resilience Plan within the scope of the Recovery and Resilience Mechanism (MRR) of the European Union (EU), framed in the Next Generation EU, for the period 2021 - 2026, within project HfPT, with reference 41.


\bibliography{main}
\addcontentsline{toc}{section}{\protect\numberline{}References}

\end{document}